# Industry-oriented Performance Measures for Design of Robot Calibration Experiment

Yier Wu[1,2], Alexandr Klimchik[1,2], Anatol Pashkevich[1,2], Stéphane Caro[2], Benoît Furet[2]

[1]*Ecole des Mines de Nantes, France, e-mail: yier.wu@mines-nantes.fr*
[2] *Institut de Recherche en Communications et Cybernétique de Nantes, France*

**Abstract**. The paper focuses on the accuracy improvement of geometric and elasto-static calibration of industrial robots. It proposes industry-oriented performance measures for the calibration experiment design. They are based on the concept of manipulator test-pose and referred to the end-effector location accuracy after application of the error compensation algorithm, which implements the identified parameters. This approach allows the users to define optimal measurement configurations for robot calibration for given work piece location and machining forces/torques. These performance measures are suitable for comparing the calibration plans for both simple and complex trajectories to be performed. The advantages of the developed techniques are illustrated by an example that deals with machining using robotic manipulator.

## 1 Introduction

Since most of the existing industrial robots employ the close-loop control on the level of the joint coordinates only, the robot accuracy highly depends on the validity of the mathematical expression used for computation of the end-effector position and orientation. These expressions include a number of parameters (geometric and elasto-static) that should be precisely identified using the data obtained from calibration experiments. However, existing methods of the optimal pose selection for calibration experiments do not take into account the particularities of many industrial applications, where the robot accuracy becomes a key factor.

In robotics, most of the previous studies concentrate on geometric calibration, which consider the effects of joint offsets and link dimensional errors [1]. Only very limited number of works address elasto-static calibration [2]. For both cases, one of the important issues is the selection of optimal measurement poses, which allow to minimize the measurement noise impact on identification accuracy. To select the optimal measurement poses, most of the authors used performance measures directly related to the covariance matrix, which should be obviously as small as possible. In particular, Khalil *et al.* used the condition number of the kinematic Jacobian [3]. Daney *et al.* applied the observability indices that are based on singular values of the covariance matrix [4, 5]. Nevertheless, these approaches, which are based on rather abstract performance measures, do not ensure the best accuracy that can be achieved after applying relevant compensation algorithm in the open-loop control of the end-effector location.

Related problem has been also studied in classical experiment design theory, which operates mainly with the linear regression models. Here, for optimal experiment design, different norms of the covariance matrix are usually formalized in T-, A-, D-optimality principles [6]. However, they are obviously not related to the primary objective of the robot calibration, its accuracy.

## 2 Problem Statement

The calibration procedure may be treated as the best fitting of the experimental data (given input variables and measured output variables) by using the geometric and/or elasto-static models. It is usually solved using the standard least-square technique, assuming that the calibration includes measurement of Cartesian coordinates that accommodate errors $\varepsilon$. These errors are assumed to be i.i.d. (independent identically distributed) random values with zero expectation and standard deviation $\sigma$. Consequently, the estimates of the identified parameters differ from their actual values. So the problem of interest is to find the measurement poses of the robotic manipulator, which ensure the lowest impact of the measurement errors (assuming that the number of poses is limited). The set of these measurement poses is defined by the corresponding joint coordinates $\mathbf{q}_i$, let us denote them as

$$\mathbf{Q} = \{\mathbf{q}_1, ..., \mathbf{q}_m\} \quad (1)$$

where $m$ is the number of experiments.

For the geometric calibration, each experiment produces two vectors $\{\Delta\mathbf{p}_i, \mathbf{q}_i\}$, which define the end-effector displacements and the corresponding joint angles. The linear relation between them can be written as

$$\Delta\mathbf{p}_i = \mathbf{J}(\mathbf{q}_i)\Delta\mathbf{\Pi} \quad (2)$$

where $\mathbf{J}(\mathbf{q}_i)$ is the Jacobian matrix that depends on manipulator configuration $\mathbf{q}_i$ and vector $\Delta\mathbf{\Pi}$ collects the unknown parameters to be identified.

For the elasto-static calibration, each experiment produces three vectors $\{\Delta\mathbf{p}_i, \mathbf{q}_i, \mathbf{F}_i\}$, where $\mathbf{F}_i$ defines the applied force and torque. In accordance with [7], the corresponding mapping can be expressed as

$$\Delta\mathbf{p}_i = \mathbf{J}(\mathbf{q}_i)\mathbf{k}_\theta \mathbf{J}^T(\mathbf{q}_i)\mathbf{F}_i \quad (3)$$

where $\mathbf{k}_\theta$ is a matrix that aggregates the unknown compliance parameters $\{k_1, ..., k_n\}$ to be identified.

In the case when both geometric and elasto-static parameters should be calibrated simultaneously, the mapping between input and output data can be expressed as

$$\Delta \mathbf{p}_i = \mathbf{J}(\mathbf{q}_i)\Delta \mathbf{\Pi} + \mathbf{J}(\mathbf{q}_i)\mathbf{k}_\theta \mathbf{J}^T(\mathbf{q}_i)\mathbf{F}_i \qquad (4)$$

From the identification theory, the unknown parameters $\Delta \mathbf{X} = \{\Delta \mathbf{\Pi}, \mathbf{k}_\theta\}$ can be obtained using the least square method, which minimizes the residuals for all experimental data. Corresponding optimization problem is formulated as

$$\sum_{i=1}^{m} \left(\Delta \mathbf{p}_i - f(\mathbf{q}_i, \Delta \mathbf{X})\right)^2 \to \min_{\Delta \mathbf{\Pi}, \mathbf{k}_\theta} \qquad (5)$$

where $f(\mathbf{q}_i, \Delta \mathbf{X})$ is defined by the right hand side of the expressions (2), (3) or (4).

In most of the related works, the calibration accuracy is evaluated with respect to statistical properties of the unknown parameter estimates obtained from Eq.(5). It can be easily proved that, under the above assumptions, the expectation of $\Delta \mathbf{X}$ is equal to the true values (i.e. the estimates are non-biased) and corresponding covariance matrix can be expressed as

$$\text{cov}(\Delta \mathbf{X}) = \sigma^2 \left(\sum_{i=1}^{m} \mathbf{B}^T(\mathbf{q}_i)\mathbf{B}(\mathbf{q}_i)\right)^{-1} \qquad (6)$$

where $\mathbf{B}(\mathbf{q}_i)$ is the transformation matrix, which can be derived from expressions (2)-(4) (required details are presented in the next section). Using this matrix, a number of approaches for design of calibration experiment were proposed (A-, T-, D-optimality). All of them are looking at minimization of the covariance matrix norm, which is equivalent to achieve the highest accuracy of the parameters. However, from practical point of view, it is necessary to achieve the best accuracy in the robot end-effector location, assuming the control is based on the obtained model. It can be proved that these two objectives are not equivalent. So, some revisions of the existing approaches in this area are required.

The goal of this work is to develop a new performance measure that ensures the highest accuracy of the robot end-effector location after error compensation. It is based on the concept of manipulator test pose (desired machining configuration), and allows essentially improving the robot accuracy at the given configuration via proper selection of measurement poses in the calibration experiments. Taking into account different application areas of the robot-based machining, the problem of defining the performance measures can be treated in accordance with two cases: (i) the robot end-effector operates without significant changes in the manipulator configuration; (ii) the robot performs machining of long and/or complex trajectories. These two cases will be address sequentially in the following sections.

## 3 Performance Measure Based on a Single Test Pose

The concept of the manipulator test pose [8] is introduced in this section for the case when the machining configuration does not have significant changes (in pick and place application and machining of small trajectories, for instance). In this particular case, it is assumed that the quality of the calibration is evaluated for the so-called test configuration. This configuration is defined by either $\{\mathbf{q}^0\}$ or $\{\mathbf{q}^0, \mathbf{F}^0\}$ for geometric and elasto-static calibration respectively, which are given by the user. For this configuration, the influence of measurement errors in the calibration experiments should be minimized in order to achieve the best end-effector location accuracy. It is defined by the mean square error of the end-effector displacements $\mathrm{E}(\delta\mathbf{p}^T\delta\mathbf{p})$, which evaluates the accuracy of errors compensation for the given test pose.

For computational convenience, the models of linear mapping that are presented in the previous section can be expressed in the following general form

$$\Delta\mathbf{p} = \mathbf{B}\cdot\Delta\mathbf{X} + \varepsilon \qquad (7)$$

where $\mathbf{B}$ is the transformation matrix and $\Delta\mathbf{X}$ collects the parameters to be identified. They vary with different cases, which are listed in Table 1.

**Table 1.** Transformation matrix and desired parameters for different cases of calibration

| Parameters to be identified | Transformation matrix $\mathbf{B}$ | Desired parameters $\Delta\mathbf{X}$ |
|---|---|---|
| (i) Geometric | $\mathbf{J}$ | $\Delta\mathbf{\Pi}$ |
| (ii) Elasto-static | $\mathbf{A}$ | $\mathbf{k}$ |
| (iii) Geometric and elasto-static | $[\mathbf{J} \quad \mathbf{A}]$ | $[\Delta\mathbf{\Pi} \quad \mathbf{k}]^T$ |

For instance, for the case of elasto-static calibration, expression (3) is rewritten using the general form (7) as

$$\Delta\mathbf{p} = \mathbf{A}\cdot\mathbf{k} \qquad (8)$$

where the $n\times 1$ vector $\mathbf{k}$ collects the desired parameters $\{k_1,...,k_n\}$, which are arranged in the $n\times n$ compliance matrix $\mathbf{k}_\theta$ in Eq.(3). Here, matrix $\mathbf{A}$ is defined by the columns of Jacobian and the external force and can be expressed as

$$\mathbf{A} = \left[\mathbf{J}^1\mathbf{J}^{1T}\mathbf{F},...,\mathbf{J}^n\mathbf{J}^{nT}\mathbf{F}\right] \qquad (9)$$

where $\mathbf{J}^n$ is the $n^{th}$ column vector of the Jacobian matrix. Similarly, expression (4) can be rewritten as

$$\Delta\mathbf{p} = \mathbf{J}\cdot\Delta\mathbf{\Pi} + \mathbf{A}\cdot\mathbf{k}. \qquad (10)$$

It can be proved that after application of the relevant compensation algorithm, the influence of the errors in the identified parameters $\delta\mathbf{X}$ and the error in the end-effector location $\delta\mathbf{p}$ can be evaluated using expression $\delta\mathbf{p} = \mathbf{B}\cdot\delta\mathbf{X}$. Besides, corresponding value of the mean square error of the robot end-effector location $\mathrm{E}(\delta\mathbf{p}^T\delta\mathbf{p})$ can be computed as $\mathrm{trace}(\delta\mathbf{p}\delta\mathbf{p}^T)$, which leads to

$$\mathrm{E}(\delta\mathbf{p}^T\delta\mathbf{p}) = \mathrm{trace}(\mathrm{E}(\mathbf{B}^0\delta\mathbf{X}\delta\mathbf{X}^T\mathbf{B}^{0T})) \qquad (11)$$

where the subscript "0" denotes the test configuration. Since $\mathrm{E}(\delta\mathbf{X}\cdot\delta\mathbf{X}^T)$ is the covariance matrix (6) of the identified parameters, the generalized performance measure can be expressed as

$$\eta^p = \sigma^2 \cdot \mathrm{trace}(\mathbf{B}^0 (\sum_{i=1}^{m} \mathbf{B}_i^T \mathbf{B}_i)^{-1} \mathbf{B}^{0T}) \tag{12}$$

where $\eta^p$ is the expected mean square errors in the end-effector location, subscript "$i$" indicates the experiment number and $m$ is the number of experiments. In more details, for different case studies described above, these performance measures are presented in Table 2.

**Table 2.** Performance measures for different parameters to be identified (single test pose)

| Parameters to be identified | Performance measures |
|---|---|
| (i) Geometric | $\eta_G^p = \sigma^2 \cdot \mathrm{trace}(\mathbf{J}^0 (\sum_{i=1}^{m} \mathbf{J}_i^T \mathbf{J}_i)^{-1} \mathbf{J}^{0T})$ |
| (ii) Elasto-static | $\eta_E^p = \sigma^2 \cdot \mathrm{trace}(\mathbf{A}^0 (\sum_{i=1}^{m} \mathbf{A}_i^T \mathbf{A}_i)^{-1} \mathbf{A}^{0T})$ |
| (iii) Geometric and elasto-static | $\eta_B^p = \sigma^2 \cdot \mathrm{trace}([\mathbf{J}^0, \mathbf{A}^0](\sum_{i=1}^{m} [\mathbf{J}_i, \mathbf{A}_i]^T [\mathbf{J}_i, \mathbf{A}_i])^{-1} [\mathbf{J}^0, \mathbf{A}^0]^T)$ |

However, it should be stressed that the proposed test-pose based approach is limited for the accuracy estimation in a single configuration. The next section deals with a more general case where changes in the manipulator configurations are significant during machining.

## 4 Performance Measure Based on a Set of Test Poses

In practice, it is often necessary to perform machining along rather long and complex trajectories (It is common for aerospace and ship building industries). This type of motions may require significant changes in the manipulator configuration and a good accuracy should be achieved for the whole machining trajectory. For this reason, it is proposed to extend the application area of the performance measure proposed in Section 3 and to minimize the maximum end-effector location errors after compensation along the trajectory. In this case, it is reasonable to apply the trajectory segmentation technique and adopt the proposed test-pose based approach for each node point. In the frame of this paper, we are not considering the problem of specifying the node points. Here, it is assumed that these points are already defined and the corresponding manipulator configurations as well as the applied forces are given. These data allow us to define a set of test pos-es $\{\mathbf{q}_j^0, \mathbf{F}_j^0 \mid j = \overline{1, s}\}$ (for geometric calibration $\mathbf{F}_j^0 = \mathbf{0}$), where $s$ is the number of node points.

For each test pose that corresponds to the node point of the trajectory, the end-effector displacement error can be computed as

$$\delta \mathbf{p}_j = \mathbf{B}_j^0 \delta \mathbf{X} \qquad (14)$$

where $\mathbf{B}_j^0$ is the transformation matrix for the configuration that is associated with the $j^{th}$ node-point. Using the technique developed in Section 3, the generalized performance measure that is based on the maximum end-effector displacement errors after compensation for all node points can be expressed as

$$\eta^t = \max_j \{ \sigma^2 \cdot \text{trace}(\mathbf{B}_j^0 (\sum_{i=1}^m \mathbf{B}_i^T \mathbf{B}_i)^{-1} \mathbf{B}_j^{0T}) \} \qquad (15)$$

where $\mathbf{B}_i$ defines the transformation matrix for the $i^{th}$ calibration experiment. In order to increase the identification accuracy, the performance measure $\eta^t$ should be minimized. Hence, it is required to solve the min-max problem. In more details, for different case studies described above, these performance measures are presented in Table 3.

**Table 3.** Performance measures for different parameters to be identified (multiple test poses)

| Parameters to be identified | Performance measures |
|---|---|
| (i) Geometric | $\eta_G^t = \max_j \{ \sigma^2 \cdot \text{trace}(\mathbf{J}_j^0 (\sum_{i=1}^m \mathbf{J}_i^T \mathbf{J}_i)^{-1} \mathbf{J}_j^{0T}) \}$ |
| (ii) Elasto-static | $\eta_E^t = \max_j \{ \sigma^2 \cdot \text{trace}(\mathbf{A}_j^0 (\sum_{i=1}^m \mathbf{A}_i^T \mathbf{A}_i)^{-1} \mathbf{A}_j^{0T}) \}$ |
| (iii) Geometric and elasto-static | $\eta_B^t = \max_j \{ \sigma^2 \text{trace}([\mathbf{J}_j^0, \mathbf{A}_j^0] (\sum_{i=1}^m [\mathbf{J}_i, \mathbf{A}_i]^T [\mathbf{J}_i, \mathbf{A}_i])^{-1} [\mathbf{J}_j^0, \mathbf{A}_j^0]^T) \}$ |

Hence, The proposed performance measures can be used both for comparing different calibration plans and as optimization functions for the experiment design. In contrast to the existing ones, the developed performance measures are directly related to the end-effector location accuracy after error compensation, which is the primary factor for industry.

## 5. Illustrative Example

Let us highlight the advantages of the proposed performance measure by an example of a 2-link planar manipulator with rigid links ($l_1 = 600\,\text{mm}$, $l_2 = 400\,\text{mm}$) and compliant joints. It is assumed that the geometric parameters are well calibrated, while the elasto-static parameters should be identified. The robot should realize a machining operation along a straight line trajectory from the point A(-600, 400) to the point B(600, 400) with a constant cutting force $\mathbf{F}^0 = \begin{bmatrix} 0 & 100\text{N} \end{bmatrix}$. This trajectory includes a set of node points that define the test poses. It is also assumed that the main source of inaccuracy is due to the measurement system used in the calibration experiments and the errors are i.i.d. (independent identically distributed) with the zero expectation and the standard deviation $\sigma = 0.1\text{mm}$. For comparison purposes, two cases are considered: one and five calibration experiments (it should be mentioned that usually the elasto-static parameters cannot be identified by one experiment, but for the considered manipulator, it gives enough data). The optimal configurations have been found both for proposed and existing performance measures (D-optimality [6] and SVD based approach (minimum singular value) [4]). For the considered cases, the deviations of the mean square errors from the target trajectory have been computed using (12) for all node points.

Relevant results are summarized in Fig. 1 and Table 4. As it shown, for the case of one experiment, the test-pose based approach ensures the accuracy of about 43% better than using SVD based approach and about 20% better comparing to the results obtained using the D-optimality. The maximum error along the whole trajectory does not overcome 0.14mm. In the case of five experiments, the test-pose based approach achieves the best accuracy of the error compensation that is 55% and 20% better comparing to the results obtained using SVD based approach and D-optimality respectively. It should be mentioned that increasing the number of experiments from one to five allowed us to improve the accuracy by a factor of 2.5 (at the same time, repeating experiments in one configuration improves the accuracy by the factor of $\sqrt{5} \approx 2.2$ only). As a result, the maximum error has been reduced to 0.055mm. Hence, the proposed performance measure that is referred to the manipulator end-effector position accuracy provides essential advantages. So, it is reasonable to use it for calibration of robot used in industrial applications.

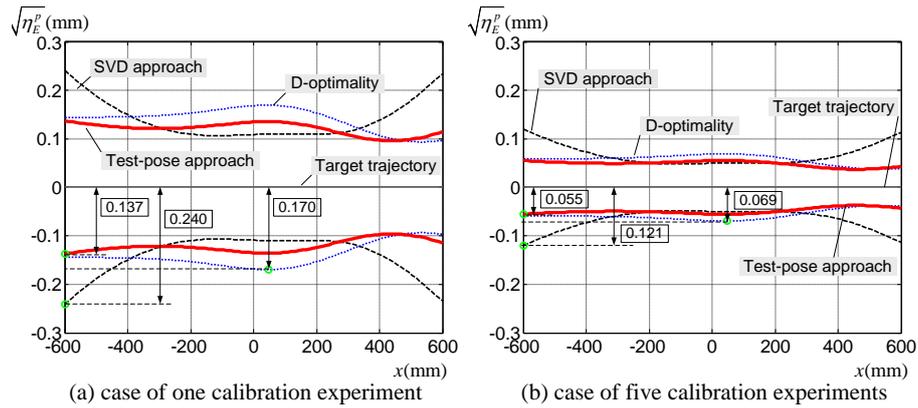

(a) case of one calibration experiment       (b) case of five calibration experiments

**Figure 1** Accuracy of the compliance errors compensation using different calibration plans

**Table 4.** Maximum deviation of square root position errors $\sqrt{\eta_E^p}$ (mm)

| Performance measure | Single experiment | 5 experiments |
|---|---|---|
| SVD based approach | 0.240 | 0.121 |
| D-optimality | 0.170 | 0.069 |
| Test-pose based approach | 0.137 | 0.055 |

## 5 Conclusions

This paper deals with the evaluation of existing approaches in the area of the calibration experiment design for robotic manipulators, taking into account particularities of some industrial applications. Particular attention is given to the identification of geometric and elasto-static parameters, whose accuracy directly influences on the quality of the robot-based machining. For this type of applications, it is proposed a new performance measure (directly related to the robot accuracy after error compensation based on the obtained parameters), which should be used for optimal design of the calibration experiments. It is shown that, for the considered case study, the proposed approach allows to improve the robot accuracy by about 20%.

**Acknowledgments** The authors would like to acknowledge the financial support of the Project ANR-2010-SEGI-003-02-COROUSSO.